\documentclass{article}

\ifdefined\pdfoutput
\pdfoutput=1
\fi

\usepackage[preprint]{neurips_2026}

\usepackage[T1]{fontenc}
\usepackage[utf8]{inputenc}
\usepackage{microtype}
\usepackage{inconsolata}

\usepackage{graphicx}
\usepackage{amsmath}
\usepackage{amssymb}
\usepackage{booktabs}
\usepackage{colortbl}
\usepackage{float}
\usepackage{url}
\usepackage[most]{tcolorbox}
\usepackage{tabularx}
\usepackage{xcolor}
\usepackage{hyperref}
\usepackage{cleveref}
\tcbuselibrary{listings,breakable}
\crefname{appendix}{Appendix}{Appendices}
\Crefname{appendix}{Appendix}{Appendices}
\hypersetup{
  pdfauthor={},
  pdftitle={Natural-Language Agent Harnesses}
}

\newcolumntype{Y}{>{\raggedright\arraybackslash}X}
\newcolumntype{C}{>{\centering\arraybackslash}X}

\newcommand{\tabref}[1]{\Cref{#1}}
\newcommand{\appref}[1]{\Cref{#1}}

\definecolor{deltapos}{HTML}{087443}
\definecolor{deltaneg}{HTML}{B42318}
\definecolor{deltaneu}{HTML}{667085}
\newlength{\rqonebenchcol}
\newlength{\rqoneharnesscol}
\newlength{\rqonetypecol}
\newcommand{\posdeltacell}[2]{#1{\scriptsize\textcolor{deltapos}{$_{#2}$}}}
\newcommand{\negdeltacell}[2]{#1{\scriptsize\textcolor{deltaneg}{$_{#2}$}}}
\newcommand{\neudeltacell}[2]{#1{\scriptsize\textcolor{deltaneu}{$_{#2}$}}}

\makeatletter
\providecommand{\@LN@col}[1]{}
\providecommand{\@LN}[2]{}
\makeatother
\newtcblisting{rqmodulebox}[1]{
  enhanced,
  breakable,
  listing only,
  colback=white,
  colframe=black,
  boxrule=0.8pt,
  arc=1mm,
  left=2mm,
  right=2mm,
  top=0.8mm,
  bottom=0.8mm,
  before skip=3pt,
  after skip=3pt,
  title=\texttt{#1},
  colbacktitle=black,
  coltitle=white,
  fonttitle=\bfseries\footnotesize,
  boxed title style={
    colframe=black,
    colback=black,
    boxrule=0.8pt,
    arc=1mm
  },
  listing engine=listings,
  listing options={
    basicstyle=\ttfamily\footnotesize,
    breaklines=true,
    columns=fullflexible,
    keepspaces=true,
    showstringspaces=false,
    aboveskip=0pt,
    belowskip=0pt
  }
}

\title{Natural-Language Agent Harnesses}

\author{
  Linyue Pan$^{1}$ \qquad Lexiao Zou$^{2}$ \qquad Shuo Guo$^{1}$ \qquad Jingchen Ni$^{1}$ \qquad Hai-Tao Zheng$^{1}$\thanks{Corresponding author.}\\
  $^{1}$Shenzhen International Graduate School, Tsinghua University \\
  $^{2}$Harbin Institute of Technology (Shenzhen)\\
  \texttt{ply24@mails.tsinghua.edu.cn} \qquad \texttt{zheng.haitao@sz.tsinghua.edu.cn}
}

\begin{document}
\maketitle

\begin{abstract}
Agent performance is strongly shaped by the surrounding harness: the external execution system around a model that organizes a task run.
Yet this logic is usually buried in tightly coupled controller code, which makes harnesses hard to inspect, compare, transfer, and ablate.
This paper asks whether the reusable design pattern of an agent harness can be represented as an executable natural-language object.
We introduce \emph{Natural-Language Agent Harnesses} (NLAHs), editable documents that describe run-level harness policy, and \emph{Intelligent Harness Runtime} (IHR), a shared runtime that interprets these documents into agent calls, handoffs, state updates, validation gates, and artifact contracts.
Across coding, terminal-use, and computer-use benchmarks, IHR-executed NLAHs achieve comparable task outcomes to code and prompted realizations, while exposing much shorter static harness policies.
Module ablations further show that explicit harness modules are analyzable.
These results suggest that agent harnesses can be turned from incidental glue around models into scientific representation objects.
\end{abstract}

\section{Introduction}

Modern language-model agents have become multi-step execution systems.
They use tools, keep state, recover from failures, validate intermediate results, and sometimes delegate work to other agents \citep{yao2023react,shinn2023reflexionlanguageagentsverbal,wang2024executablecodeactionselicit,fourney2024magenticonegeneralistmultiagentsolving,anthropic2024buildingeffectiveagents}.
These behaviors are organized by an external \emph{harness}, which can have large effects on measured performance \citep{langchain2026improvingdeepagents,langchain2026anatomyagentharness,bui2026buildingeffectiveaicoding}.
Similar concerns appear in recent work on agent scaffolds, workflow generation, long-context execution, multi-agent orchestration, and tool-using agents \citep{openai2026harnessengineering,anthropic2025effectiveharnesses,anthropic2025multiagentresearchsystem,ding2026octobenchbenchmarkingscaffoldawareinstruction,liu-etal-2024-lost,chroma2025contextrot,sun2025scalinglonghorizonllmagent}.

\begin{figure}[t]
\centering
\includegraphics[width=\textwidth]{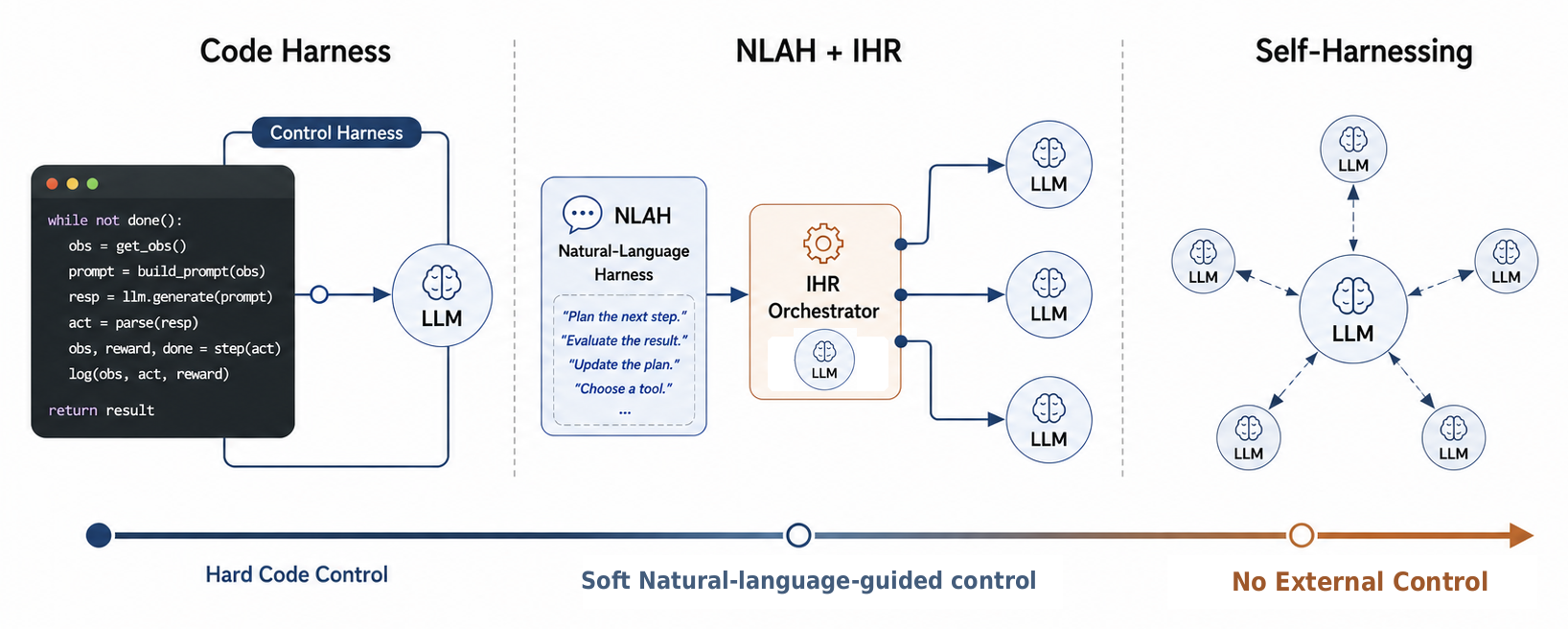}
\caption{\textbf{Three ways to control an agent run.}
The spectrum ranges from restrictive harnessing to no external harness, or self-harnessing.
Code harnesses impose hard external control on a model through program logic.
NLAH+IHR, the design point studied in this paper, moves the harness policy into readable natural language while a shared runtime executes that policy through child-agent calls.
Self-harnessing is a possible future design in which a controller model directly harnesses other models without any external harness.}
\label{fig:harnessing-spectrum}
\end{figure}

The problem is that harnesses are usually not represented as clean research objects.
A code harness may mix prompts, tool adapters, parser rules, validation scripts, artifact paths, retry logic, context policy, and benchmark-specific assumptions in one controller bundle.
As a result, a seemingly small harness change can also change call boundaries, tool mediation, state carriers, validation gates, and stopping semantics.
This makes harnesses hard to inspect, port, compare, and ablate, even though the harness pattern itself is often the reusable part of the system.

This paper studies whether a harness pattern can be externalized as executable natural language.
We propose \emph{Natural-Language Agent Harnesses} (NLAHs), which write run-level harness policy as editable text, and \emph{Intelligent Harness Runtime} (IHR), a shared runtime that executes this policy through agent calls.
The key separation is simple: natural language carries the harness policy, while code and the runtime carry exact mechanisms such as tool execution, parsing, sandboxing, and logging.

We evaluate this idea with three connected questions.
First, can IHR-executed NLAHs control real agent runs while preserving task performance comparable to code and prompted realizations?
Second, do IHR-executed NLAHs materialize the intended harness mechanisms beyond using the same text as ordinary prompting?
Third, once harness policy is explicit, can individual modules such as file-backed state, verifier separation, self-evolution, and multi-candidate search be analyzed as module-level interventions?
Across coding, terminal-use, and computer-use benchmarks, our results show that NLAHs are executable and compact, that they leave measurable behavioral traces, and that module-level gains depend on whether a module aligns intermediate control with the benchmark's acceptance condition.

Our contributions are as follows:

\begin{itemize}
\item We introduce NLAHs as explicit natural-language representations of agent harness patterns, distinct from both runtime policy and deterministic code hooks.
\item We introduce IHR, a shared in-loop runtime that turns NLAHs into auditable agent calls, handoffs, state updates, validation gates, and artifact contracts.
\item We explore the boundary between natural language and code in agent-harness systems, making a first step toward substantially broadening the scope of natural language from local instructions to harness-level strategy.
\item We provide controlled evidence across three benchmark families that NLAHs can shape agent behavior with comparable task outcomes, expose concise static harness policies, and support module-level analysis.
\end{itemize}

\section{Preliminaries}
\label{sec:preliminaries}

A \emph{model} is a callable learned function from context $c$ to output $y$, where the context may include text, images, or video:
\[
y = \operatorname{LM}_m(c).
\]

An \emph{agent} is a system that wraps one or more model calls with external interaction.
An agent receives a task, maintains some execution state, observes feedback from tools or environments, and decides whether to continue acting, ask for information, validate progress, or stop.
Thus, an agent is a model-centered execution process that can include multiple model calls and external actions.
A single model call is a degenerate special case of an agent call, where the agent is allowed to call the model only once for a one-shot answer and performs no external action.
In this paper, the atomic unit of harness execution is therefore an \emph{agent call}.
This choice lets NLAH describe harness behavior at the level where prompts, tools, state, validation, and delegation actually operate.

A \emph{harness} is the external execution system around a model in an agent.
It turns a base model into an agent that can act over real tasks by deciding what the model sees, what tools it may call, where state is stored, how observations are returned, when validation runs, how failures are recovered, when execution may stop, and how one or more model or agent calls are organized.
\emph{Harness engineering} is the practice of designing, implementing, adapting, debugging, and evaluating agent harnesses.
\appref{app:harness-aspects} summarizes descriptions of eleven main aspects of harness engineering.
These include agent loops, tool design and documentation, context engineering, filesystem and workspace management, memory and state, validation and stopping conditions, safety permissions and sandboxing, runtime defaults, observability and replay, retry and recovery, and budget control.

\section{Methodology}

\begin{figure}[t]
\centering
\includegraphics[width=\textwidth]{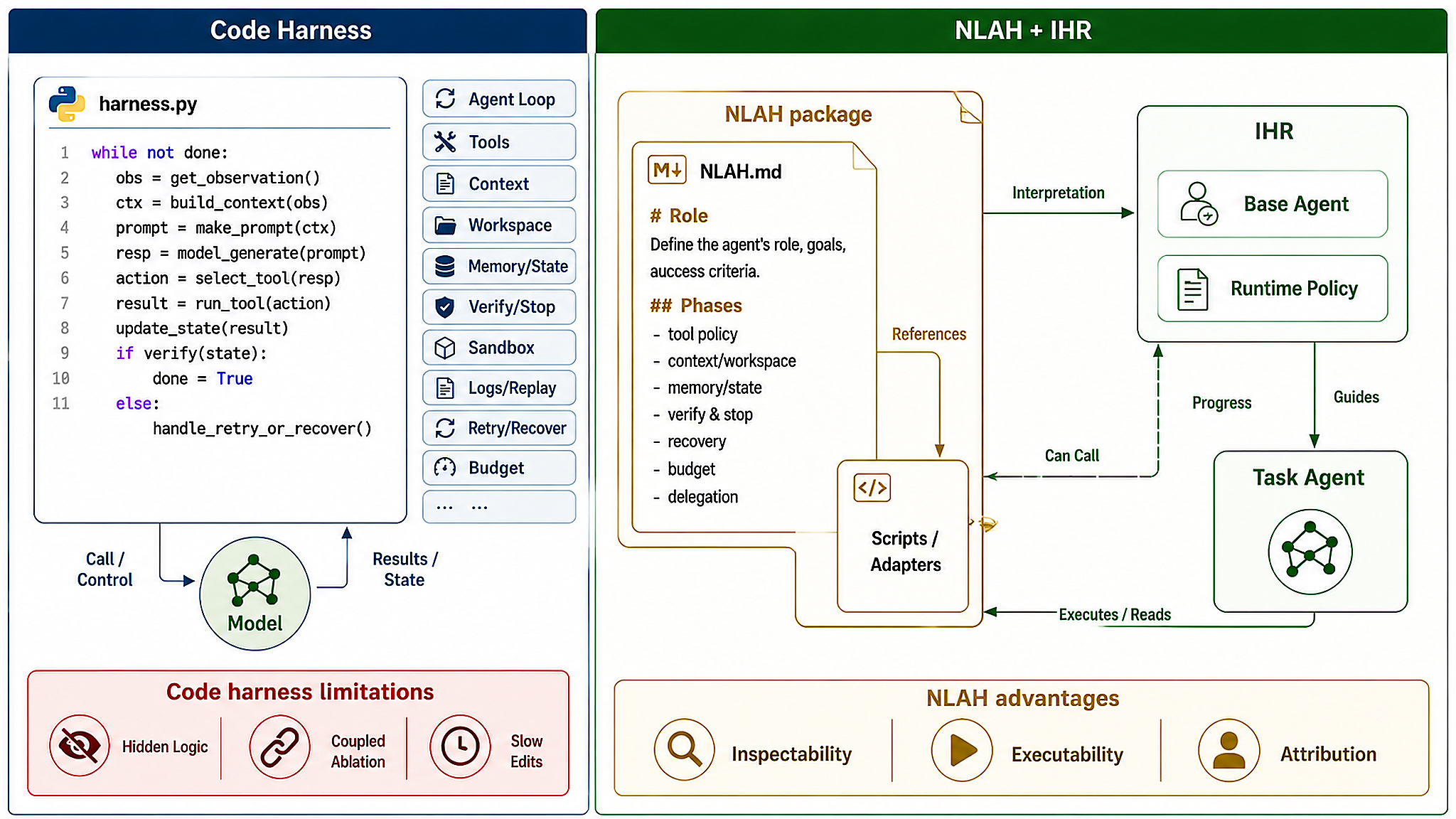}
\caption{\textbf{NLAH+IHR framework.}
A native code harness mixes policy and mechanism inside controller code.
NLAH+IHR separates them: the NLAH stores readable harness policy, IHR provides shared execution semantics, and scripts or adapters handle exact operations such as tools, tests, parsers, and validators.}
\label{fig:framework-overview}
\end{figure}

Inspired by reusable natural-language carriers such as \texttt{AGENTS.md}, \texttt{CLAUDE.md}, and \texttt{SKILL.md}, we consider extending natural-language documents from simple tool or workflow descriptions to broader harness-level strategies.

\subsection{NLAHs and IHR}
\label{sec:nlahs-ihr}
An NLAH+IHR system has four layers.
The first layer is the \emph{base agent}: a code-form minimal executable substrate.
In our setting, the base agent is only an LLM loop: it can call a model, but the only external tool exposed to the model is a terminal.
Through the terminal, the base agent can read and write files, run processes, record events, and launch child agents when needed.
Launching a child agent does not require a separate dedicated tool: the base agent can use the terminal to start a new instance of itself and pass that instance a child task packet.
The second layer is the \emph{runtime policy}: a fixed instruction that turns the base agent into IHR by defining how it should interpret and execute harness documents.
The third layer is the \emph{NLAH}: the natural-language policy document that describes the stages, roles, state rules, verification rules, recovery rules, and stopping conditions of a task run.
The fourth layer is the set of \emph{scripts and adapters}: deterministic code used for exact operations such as running tests, parsing results, calling benchmark tools, or checking artifacts.

This separation is the main design choice.
The base agent and adapters provide the machine interface.
The runtime policy provides shared execution semantics.
The NLAH provides the per-harness policy.

IHR is intentionally thin, containing the base agent and the text-form runtime policy.
It uses the base agent as an orchestrator guided by the runtime policy and delegates substantive task work to child agents.
For a nominally single-agent harness, IHR still realizes the run as a parent orchestrator plus one executor child, so the boundary between harness control and task execution remains visible.
For multi-role or multi-branch harnesses, IHR launches separate child agents, passes each agent only the intended task packet, supervises handoff, and records the resulting behavior.
Thus, IHR is not a large bespoke controller for one benchmark; it is a shared runtime that gives natural-language harness policy a common execution substrate.

The NLAH is the part that changes from one harness to another.
It specifies what the run should do and leaves low-level operations to the runtime, tools, and hooks.
For example, an NLAH can state when to create a task state file, when to ask a verifier to inspect a patch, what evidence must be preserved before answering, when a retry is allowed, and what condition closes the run.
The runtime then instantiates these clauses through model calls, child-agent messages, tool calls, files, and deterministic hooks.
NLAHs extend beyond ordinary prompts: they describe the lifecycle of a task run and cover subsequent multi-step execution.

We keep deterministic code where precision matters.
Tests, parsers, sandboxing, benchmark adapters, and artifact validators remain in code because they require exact and reproducible behavior.
Natural language is used for policy: task decomposition, role contracts, evidence discipline, retry logic, state handoff, and validation strategy.
This division of labor avoids the overclaim that natural language can replace all controller code, while still moving the most inspectable part of harness design out of opaque implementation logic.
\appref{app:nlah-expressivity} gives the natural-language/code boundary and maps code-harness aspects to the IHR+NLAH carriers.

\subsection{Notes on writing NLAHs}
\label{sec:writing-nlahs}
In our experiments, an NLAH is a compact policy document that makes the harness decisions explicit.
We found the following writing principles useful.

\paragraph{State the task contract first.}
An NLAH should begin by defining the input, the expected output, the allowed tools or artifacts, and the condition under which the run is complete.
This prevents later sections from becoming vague advice.
For coding tasks, the contract may specify patch location, test evidence, and final answer format.
For computer-use tasks, it may specify the target application state, allowed interaction channels, and completion evidence.

\paragraph{Separate stages from mechanisms.}
The NLAH should name the stages of the run---for example, inspect, plan, edit, verify, recover, and finalize---but it should not reimplement every low-level tool operation in prose.
Low-level operations are better handled by scripts, adapters, and runtime hooks.
The NLAH should instead define when those mechanisms are used and what evidence they must produce.

\paragraph{Make state and evidence explicit.}
Long-horizon agents fail when useful intermediate information is lost or when a final answer is produced without auditable evidence.
A readable NLAH should therefore specify where state is stored, which artifacts must be reopened by later agents, what evidence supports a claim, and which files or logs close the run.
This is especially important for file-backed state, verifier modules, and evidence-backed answering.

\paragraph{Write module boundaries so they can be ablated.}
A module is useful for research only if it can be removed or changed without silently changing the rest of the harness.
NLAH sections should therefore use clear names for modules such as verifier, self-evolution, multi-candidate search, context compression, or markdown memory.
This lets us ask whether a module changes task outcomes, process metrics, or solved-set composition under a shared runtime.

\paragraph{Prefer simple and enforceable language.}
NLAHs should use short clauses, concrete conditions, and explicit artifacts.
Phrases such as ``be careful,'' ``think deeply,'' or ``act like an expert'' are weak harness policy because they do not define observable behavior.
By contrast, clauses such as ``write a state file before delegating,'' ``run the verifier only after producing a candidate patch,'' or ``do not finalize without evidence from the target file'' are easier for IHR to execute and easier for researchers to audit.
\appref{app:nlah-expressivity} gives the detailed expressivity boundary.

\section{Experimental Design}

\subsection{Research questions}
We evaluate whether harness pattern logic can be compared across implementation media, audited through harness mechanisms, and analyzed as explicit modules.

\begin{itemize}
\item \textbf{RQ1 (Harness Realization).} Can NLAHs shape observable agent behavior while maintaining comparable task outcomes, and how does this control compare with native code harnesses and prompted NLAHs?
\item \textbf{RQ2 (Harness Mechanism Realization).} Do IHR-executed NLAHs preserve and materialize intended harness mechanisms, such as workflow structure, contract enforcement, tool use, recovery, and information handoff?
\item \textbf{RQ3 (Module Ablation).} Once harness modules are expressed in natural language, can they be cleanly ablated and analyzed at the module level?
\end{itemize}

\subsection{Harness realizations}
RQ1 compares three realizations of the same harness idea, ordered by how directly the harness can control execution.
\begin{itemize}
\item \textbf{Code Harness} denotes the original code implementation of the studied agent harness family: the controller code, workflow scripts, framework defaults, and tool adapters that realize the harness before it is represented as an NLAH.
It provides the strongest and most deterministic control, but its policy is interleaved with implementation details.
\item \textbf{Prompted NLAH} denotes the same NLAH content provided as ordinary prompt or instruction text to the Codex CLI agent, without IHR's shared runtime charter and execution semantics.
It tests how much control is available when natural language is only a passive instruction carrier.
\item \textbf{IHR-executed NLAH} denotes NLAH interpreted and executed by IHR, with explicit runtime semantics for child lifecycle, artifact and state handling, contract gates, and stopping.
It gives up the hard determinism of a code harness, but gives the natural-language policy an execution substrate that can materialize roles, handoffs, state, and verification boundaries.
\end{itemize}

\subsection{Benchmarks and harness families}
We evaluate on three representative benchmark families that require multi-step control, tool use, durable state accumulation, and verification or evidence management.

\paragraph{Coding.}
SWE-bench Verified evaluates repository-grounded issue resolution; the main metric is issue resolution rate \citep{jimenez2024swebench,chowdhury2024swebenchverified}.
We study coding harness families including Live-SWE-Agent \citep{xia2025livesweagentsoftwareengineeringagentssoftwareengineeringagents}.

\paragraph{Terminal-use capability.}
Terminal-Bench~2.0 (TB2) evaluates long-horizon command-line tasks in Linux environments; the main metric is task success \citep{merrill2026terminalbenchbenchmarkingagentshard}.
Meta-Harness is an agent-driven technique that automatically debugs and optimizes executable code harnesses \citep{lee2026metaharnessendtoendoptimizationmodelmodel}.
We study MHTBA, the state-of-the-art terminal-use code harness produced by Meta-Harness for Terminal-Bench~2.0 with Claude Opus 4.6 \citep{metaharness2026tbench2artifact}.

\paragraph{Computer use.}
OSWorld evaluates computer-use behavior grounded in real desktop environments; the main metric is task success rate \citep{NEURIPS2024_5d413e48}.
For OSWorld, we report a SeeAct-style GUI harness family \citep{zheng2024gpt4visiongeneralistwebagentgrounded}.

\subsection{Experimental setup}
All experiments use the same IHR instantiation: Codex CLI version \texttt{0.123.0}, model \texttt{gpt-5.4-mini} \citep{openai2026gpt54mininano}, and reasoning effort \texttt{xhigh}.
Runs execute on Ubuntu 24.04 servers with 64 CPU cores and 251~GiB of memory.
To improve reproducibility and sandbox safety, all runs are executed in Docker containers.
Per-task container caps are 32 vCPUs, 84~GiB memory, and 40~GiB storage.

\section{Results}

\subsection{RQ1: Harness realization}
\begin{table}[t]
\centering
\small
\setlength{\tabcolsep}{1.6pt}
\caption{\textbf{RQ1: NLAH execution preserves competitive task performance while exposing process costs.}
Perf. is the benchmark primary percentage metric.
Code denotes the native code harness, Prompt denotes the same NLAH text used as ordinary instructions, and NLAH denotes IHR-executed NLAH.
Token and call metrics report the observable process cost of each realization.}
\label{tab:rq1}
\begin{tabularx}{\linewidth}{@{}>{\raggedright\arraybackslash}m{\rqonebenchcol}
>{\raggedright\arraybackslash}m{\rqoneharnesscol}
>{\raggedright\arraybackslash}m{\rqonetypecol}
*{6}{>{\raggedleft\arraybackslash}X}@{}}
\toprule
Benchmark & Harness & Type & Perf. & \shortstack[c]{LLM\\Calls} & \shortstack[c]{Tool\\Calls} & \shortstack[c]{Pr.\\Tok.} & \shortstack[c]{Comp.\\Tok.} & \shortstack[c]{Run\\time\\(min)} \\
\midrule
\mbox{SWE Verified} & \mbox{Live-SWE} & Code & 67.00 & 23.30 & 17.70 & 283.60k & 3.50k & 28.90 \\
& & Prompt & 77.00 & 36.40 & 48.00 & 2.20M & 27.50k & 5.70 \\
& & NLAH & 73.00 & 41.00 & 63.40 & 2.20M & 32.30k & 6.10 \\
TB2 & MHTBA & Code & 36.00 & 223.20 & 122.90 & 10.40M & 17.50k & 19.50 \\
& & Prompt & 57.30 & 41.50 & 48.00 & 3.10M & 51.80k & 11.10 \\
& & NLAH & 53.90 & 56.40 & 78.00 & 4.20M & 74.80k & 13.50 \\
OSWorld & SeeAct & Code & 47.10 & 23.30 & 47.80 & 1.40M & 8.90k & 9.00 \\
& & Prompt & 47.90 & 35.30 & 39.20 & 1.10M & 12.30k & 4.90 \\
& & NLAH & 46.30 & 40.90 & 48.60 & 1.10M & 13.60k & 5.50 \\
\bottomrule
\end{tabularx}
\end{table}

RQ1 asks whether harness policy can be moved from code into an NLAH without losing the ability to control real agent runs.
We compare three realizations: the native code harness, the same NLAH used as ordinary instructions, and the same NLAH executed by IHR.
This design separates two questions: whether natural-language harness policy is expressive enough, and whether a shared runtime gives that policy stronger execution semantics than prompting alone.
The results are shown in \tabref{tab:rq1}.

\paragraph{IHR-executed NLAHs are operationally viable.}
Across the audited settings, NLAHs achieve task performance in the same regime as the corresponding code harnesses.
On Live-SWE, IHR-executed NLAH reaches 73.0, above the native code harness at 67.0 and close to the prompted NLAH at 77.0.
On OSWorld, NLAH reaches 46.3, essentially matching the code harness at 47.1.
On MHTBA, NLAH reaches 53.9, below the prompted version at 57.3 but far above the native code realization at 36.0.
A detailed analysis of the MHTBA code artifact's TB2 portability is given in \appref{app:mhtba-portability}.
These results support the central feasibility claim: NLAH and IHR together can drive real multi-step agent behavior.

\paragraph{The cost profile reflects prototype-runtime engineering overhead.}
NLAHs often use more model calls, tool calls, or tokens than code harnesses.
This is expected in the current implementation because IHR is built on a general agent substrate and uses natural-language orchestration, which adds overhead relative to a hand-specialized controller.
The important point is that this cost does not destroy task performance.
In several cases, the added autonomy lets the model choose action granularity more flexibly than a rigid controller, which helps explain why Live-SWE NLAH is both competitive and much faster in wall-clock time than the native code harness.
Thus, the current cost profile should be read as an engineering target; it does not show that the representation is unusable.

\paragraph{NLAHs expose the policy layer that code harnesses hide.}
The conciseness audit in \tabref{tab:rq1-conciseness} shows the strongest representation-level result.
For Live-SWE, the readable harness policy is reduced from 60.1k tokens of code materials to a 2.9k-token NLAH.
For MHTBA, it is reduced from 10.5k to 0.8k tokens.
This means the high-level policy---state handling, validation, recovery, candidate search, and completion gates---is separated from deterministic mechanisms and becomes directly inspectable.
That separation is what enables mechanism-level auditing in RQ2 and module-level ablation in RQ3.

\paragraph{Behavior is flexible but still policy-guided.}
The OSWorld cases illustrate why NLAH is best viewed as a policy layer operating at the level of goals, evidence, and gates.
The NLAH preserves staged observation, action selection, recovery, and completion checking, but it may choose a different concrete route when that route satisfies the same completion contract.
For example, GUI-oriented tasks can sometimes be completed through shell commands, file edits, or package-level operations that provide clearer evidence.
This flexible routing preserves harness control and reflects the intended benefit of expressing policy at the level of goals, evidence, and gates; the policy need not prescribe every action primitive.

\paragraph{Takeaway.}
RQ1 supports the first claim of the paper: harness policy can be externalized into compact natural language and executed by a shared runtime while preserving competitive task outcomes.
The main remaining gap is engineering efficiency: reducing handoff loss, fixed context overhead, and redundant orchestration calls in the prototype runtime.

\begin{table}[t]
\centering
\small
\caption{\textbf{RQ1: NLAHs expose the reusable harness policy in fewer static materials.}
Counts include audited static NLAH files and corresponding code-harness implementation materials, not runtime prompts or generated logs.}
\label{tab:rq1-conciseness}
\begin{tabular*}{\linewidth}{@{\extracolsep{\fill}}llrrrr@{}}
\toprule
\multicolumn{1}{@{}l}{Benchmark} & Harness & \multicolumn{2}{c}{Token} & \multicolumn{2}{c@{}}{Files} \\
\cmidrule(lr){3-4}\cmidrule(l){5-6}
& & Code & NLAH & Code & NLAH \\
\midrule
SWE Verified & Live-SWE & 60.10k & 2.90k & 68.00 & 3.00 \\
TB2 & MHTBA & 10.50k & 0.80k & 3.00 & 1.00 \\
OSWorld & SeeAct & 47.50k & 1.40k & 5.00 & 1.00 \\
\bottomrule
\end{tabular*}
\end{table}

\subsection{RQ2: Harness mechanism realization}
RQ2 asks whether IHR-executed NLAHs materialize the intended harness mechanisms in addition to matching task scores.
We define and use new pattern-preservation and harness-engineering metrics for the settings where logs expose the required event structure.
Because NLAHs deliberately operate at the level of policy, contracts, and gates, the audit focuses on whether expected mechanisms such as workflow structure, stage coverage, tool use, contract enforcement, recovery, and handoff appear in the run.

\begin{table}[t]
\centering
\scriptsize
\setlength{\tabcolsep}{2pt}
\caption{\textbf{RQ2: NLAHs preserve recognizable harness-pattern structure.}
Except for Verification Signals, the Code row is the reference harness and therefore does not receive one-way similarity scores.}
\label{tab:rq2-pattern-preservation}
\resizebox{\textwidth}{!}{%
\begin{tabular}{@{}lllrrrrrrrr@{}}
\toprule
Benchmark & Harness & Type & \shortstack[c]{Verification\\Signals} & \shortstack[c]{Prompt\\Contract} & \shortstack[c]{Tool\\Surface} & \shortstack[c]{Workflow\\Pres.} & \shortstack[c]{Stage\\Cov.} & \shortstack[c]{Ordered\\Workflow} & \shortstack[c]{Context\\Boundary} & \shortstack[c]{Model\\Match} \\
\midrule
SWE Verified & Live-SWE & Code & 3.99 & - & - & - & - & - & - & - \\
SWE Verified & Live-SWE & Prompt & 6.51 & 0.89 & 0.82 & 0.70 & 0.75 & 0.74 & 1.00 & 1.00 \\
SWE Verified & Live-SWE & NLAH & 9.89 & 0.81 & 0.87 & 0.67 & 0.82 & 0.78 & 0.76 & 0.76 \\
Terminal-Bench 2.0 & MHTBA & Code & 45.05 & - & - & - & - & - & - & - \\
Terminal-Bench 2.0 & MHTBA & Prompt & 13.18 & 1.00 & 0.81 & 0.64 & 0.57 & 0.53 & 1.00 & 0.99 \\
Terminal-Bench 2.0 & MHTBA & NLAH & 22.82 & 0.84 & 0.80 & 0.63 & 0.57 & 0.54 & 0.81 & 0.55 \\
\bottomrule
\end{tabular}}
\end{table}

\paragraph{NLAHs preserve recognizable workflow structure.}
\tabref{tab:rq2-pattern-preservation} shows that NLAH runs keep nontrivial prompt-contract, tool-surface, workflow-preservation, stage-coverage, and ordered-workflow scores relative to the reference harnesses.
On Live-SWE, NLAH raises Verification Signals to 9.890 and improves Stage Coverage and Ordered Workflow over prompted execution.
On MHTBA, NLAH has nearly the same Workflow Preservation as Prompt, slightly higher Stage Coverage and Ordered Workflow, and lower Context Boundary and Model Match because parent-child execution changes the topology and distributes work across parent and child contexts.
These metrics support the same qualitative point as RQ1's behavioral discussion: NLAH execution is policy-guided execution.

\begin{table}[t]
\centering
\scriptsize
\setlength{\tabcolsep}{3pt}
\caption{\textbf{RQ2: NLAHs instantiate harness-engineering mechanisms.}
For Prompt, Orchestration Reliability and Information Handoff Recall use direct-context variants; for NLAH, they use parent-child handoff variants.}
\label{tab:rq2-harness-engineering}
\resizebox{\textwidth}{!}{%
\begin{tabular}{@{}lllrrrrrr@{}}
\toprule
Benchmark & Harness & Type & \shortstack[c]{Artifact\\Contract} & \shortstack[c]{Tool Call\\Success} & \shortstack[c]{Failed Tool\\Continuation} & \shortstack[c]{Cached Token\\Ratio} & \shortstack[c]{Orchestration\\Reliability} & \shortstack[c]{Information\\Handoff Recall} \\
\midrule
SWE Verified & Live-SWE & Code & 0.99 & 0.88 & 0.95 & 0.71 & NA & NA \\
SWE Verified & Live-SWE & Prompt & 0.99 & 0.93 & 0.98 & 0.96 & 1.00 & 1.00 \\
SWE Verified & Live-SWE & NLAH & 1.00 & 0.93 & 0.99 & 0.94 & 0.83 & 0.32 \\
Terminal-Bench 2.0 & MHTBA & Code & NA & 0.95 & 0.79 & 0.00 & NA & NA \\
Terminal-Bench 2.0 & MHTBA & Prompt & 1.00 & 0.92 & 1.00 & 0.96 & 0.99 & 1.00 \\
Terminal-Bench 2.0 & MHTBA & NLAH & 0.96 & 0.93 & 1.00 & 0.94 & 0.85 & 0.55 \\
\bottomrule
\end{tabular}}
\end{table}

\paragraph{The strongest mechanism evidence appears in contracts, tools, and recovery.}
\tabref{tab:rq2-harness-engineering} shows high artifact-contract compliance, tool-call success, and continuation after failed tool calls for IHR-executed NLAHs.
On Live-SWE, NLAH reaches 1.000 Artifact Contract, 0.933 Tool Call Success, and 0.992 Failed Tool Continuation.
On MHTBA, the corresponding values are 0.955, 0.928, and 0.995.
These numbers indicate that IHR turns instructions from text into observable artifacts, tool-mediated execution, and recovery behavior.

\paragraph{The main mechanism weakness is handoff.}
The same table also identifies the main runtime bottleneck.
NLAH Orchestration Reliability is lower than Prompt on both Live-SWE and MHTBA, and Information Handoff Recall drops from the direct-context Prompt setting to 0.322 and 0.553 under parent-child execution.
This weakness is consistent with the cost profile in RQ1: the prototype runtime already materializes harness mechanisms, but loses information across boundaries that ordinary prompting does not create.

\paragraph{Takeaway.}
RQ2 separates the mechanism claim from RQ1's outcome claim.
The same runs show that NLAH+IHR remains competitive on task outcomes and also produces auditable workflow, contract, verification, tool-use, recovery, and handoff signals.
The main remaining gap is therefore more specific than generic overhead: IHR needs better handoff and orchestration reliability.

\subsection{RQ3: Module ablation}

\begin{table}[t]
\centering
\small
\setlength{\tabcolsep}{3pt}
\caption{\textbf{RQ3: Explicit NLAH modules can be ablated under a shared runtime.}
Each row adds one module to a benchmark-specific Basic condition.
Perf. is the main benchmark metric, and Agent Calls measures changes in execution topology.
Values should be compared within the same benchmark column, not across benchmark families.}
\label{tab:rq3}
\begin{tabularx}{\textwidth}{@{}>{\raggedright\arraybackslash}X
*{4}{>{\centering\arraybackslash}p{0.14\textwidth}}@{}}
\toprule
Setting & \multicolumn{2}{c}{SWE Verified} & \multicolumn{2}{c}{OSWorld} \\
\cmidrule(lr){2-3}\cmidrule(l){4-5}
& Perf. & \shortstack[c]{Agent\\Calls} & Perf. & \shortstack[c]{Agent\\Calls} \\
\midrule
Basic & 73.00 & 1.10 & 44.40 & 1.08 \\
+ File-backed state & \posdeltacell{75.60}{+2.60} & \neudeltacell{1.10}{0.00} & \posdeltacell{58.30}{+13.90} & \neudeltacell{1.11}{+0.03} \\
+ Evidence-backed answering & \posdeltacell{75.80}{+2.80} & \neudeltacell{1.20}{+0.10} & \posdeltacell{47.20}{+2.80} & \neudeltacell{1.06}{-0.03} \\
+ Verifier & \posdeltacell{73.20}{+0.20} & \neudeltacell{2.30}{+1.20} & \posdeltacell{52.80}{+8.40} & \neudeltacell{1.42}{+0.33} \\
+ Self-evolution & \posdeltacell{78.80}{+5.80} & \neudeltacell{1.20}{+0.10} & \posdeltacell{52.80}{+8.40} & \neudeltacell{1.19}{+0.11} \\
+ Multi-candidate search & \negdeltacell{71.40}{-1.60} & \neudeltacell{5.70}{+4.60} & \posdeltacell{47.20}{+2.80} & \neudeltacell{1.33}{+0.25} \\
+ Dynamic orchestration & \posdeltacell{74.60}{+1.60} & \neudeltacell{1.60}{+0.50} & \posdeltacell{47.20}{+2.80} & \neudeltacell{1.14}{+0.06} \\
+ Context compression & \negdeltacell{72.00}{-1.00} & \neudeltacell{2.20}{+1.10} & \negdeltacell{36.10}{-8.30} & \neudeltacell{1.22}{+0.14} \\
+ Markdown memory & \negdeltacell{70.20}{-2.80} & \neudeltacell{1.30}{+0.20} & \posdeltacell{50.00}{+5.60} & \neudeltacell{1.54}{+0.46} \\
\bottomrule
\end{tabularx}
\end{table}

RQ3 asks whether explicit NLAH modules support meaningful intervention under a shared runtime.
We analyze \tabref{tab:rq3} from a global perspective: which module families help across benchmarks, which ones mainly change process shape, and which ones add cost or branching without improving the path to benchmark acceptance.
The discussion compares modules within each reported benchmark; averaging scores across benchmark families would be inappropriate.
We keep the two most consequential observations in the main text and move the remaining module-specific discussion to \appref{app:rq3-additional}.

\paragraph{The strongest modules tighten state and acceptance discipline.}
Two modules stand out.
\emph{File-backed state} improves both benchmarks, from 73.0 to 75.6 on SWE and from 44.4 to 58.3 on OSWorld.
\emph{Self-evolution} is even stronger on the solve loop itself, reaching 78.8 on SWE and 52.8 on OSWorld.
\emph{Evidence-backed answering} is also consistently positive, though more modestly, with gains of +2.8 on both benchmarks.
The common pattern is important: the modules that help most make the acceptance path cleaner by preserving state, forcing explicit evidence, or sharpening the retry decision.

\paragraph{Extra branching is not the same as better control.}
\emph{Multi-candidate search} produces the clearest topology change.
Agent Calls jump from 1.1 to 5.7 on SWE and from 1.083 to 1.333 on OSWorld, but this extra branching yields only +2.8 on OSWorld and a drop from 73.0 to 71.4 on SWE.
This is a useful negative result.
Explicit branching does change the search behavior, but under the current runtime and budget it is too expensive and too infrastructure-sensitive to dominate simpler modules.
More search is not automatically better harness design.

\paragraph{Global takeaway.}
RQ3 therefore supports a clear paper-level conclusion.
Explicit NLAH modules are useful when they shorten the path from intermediate work to auditable evidence and final benchmark acceptance.
They are less useful when they mainly add local process layers, extra branching, or compressed summaries whose notion of success can drift away from the evaluator.
This is exactly the kind of conclusion that is hard to reach when harness logic stays buried inside code: once the modules become explicit, we can see whether they help and \emph{how} they help.

\section{Related Work}

\paragraph{Agent harnesses and scaffold-aware evaluation.}
Recent agent systems show that performance depends on the execution scaffold around the model, including tools, feedback loops, state, validation, and workflow structure \citep{yao2023react,lewis2021retrievalaugmentedgenerationknowledgeintensivenlp,shinn2023reflexionlanguageagentsverbal,fourney2024magenticonegeneralistmultiagentsolving,muennighoff2025s1simpletesttimescaling,anthropic2024buildingeffectiveagents}.
Code-harness synthesis, scaffold-aware benchmarks, agent graph compilation, and multi-agent routing or orchestration further make this dependence explicit \citep{lou2026autoharnessimprovingllmagentsimprovingllmagents,chivukula2025agintagenticgraphcompilation,ding2026octobenchbenchmarkingscaffoldawareinstruction,an2025amobenchlargelanguagemodels,zhan2026mathsmithextremelyhardmathematical,zhan20263viewsensespatialmentalperspective,yang2026ossymphonyholisticframeworkrobust,wang-etal-2025-anymac,wang-etal-2025-agentdropout,yue-etal-2025-masrouter,ke2026masorchestraunderstandingimprovingmultiagent,costa2026agentspawnadaptivemultiagentcollaboration}.
These works motivate our setting.
Our focus is whether the harness policy itself can be externalized as editable natural language and executed under a shared runtime.

\paragraph{Natural-language instruction carriers.}
Files such as prompts, \texttt{AGENTS.md}, \texttt{CLAUDE.md}, AgentSkills, and related skill bundles show that operational knowledge can be packaged as reusable text and attached to agent runs \citep{agentsmd2026format,agentskillsio2026home}.
Recent skill and memory work extends this idea by learning, evolving, storing, and transferring reusable procedures \citep{hao2026recreatereasoningcreatingdomain,ye2026metacontextengineeringagentic,mi2026procmemlearningreusableprocedural,zhang2026memskilllearningevolvingmemory,li2026organizingorchestratingbenchmarkingagent,li2026singleagentskillsreplacemultiagent,li2026skillsbenchbenchmarkingagentskills,chen2026skillcraftllmagentslearn,pinchbench2026pinchbench,xia2026skillrlevolvingagentsrecursive}.
NLAHs are related but operate at a different level.
NLAHs operate at the run-level harness-policy layer, specifying roles, call boundaries, state carriers, evidence gates, recovery rules, and stopping criteria.

\paragraph{Natural language as programs, workflows, and constraints.}
Prompt programming, promptware, and language-model programming frameworks such as LMQL, DSPy, APPL, and SGLang treat prompts and LLM calls as programmable objects \citep{10.1145/3729342,cheng2025sharingstatepromptsprograms,chen2026promptwareengineeringsoftwareengineering,10.1145/3591300,ICLR2024_f1cf02ce,dong-etal-2025-appl,10.5555/3737916.3739916}.
Other work compiles natural language into workflows, graphs, runtime constraints, or executable specifications \citep{li2024autoflowautomatedworkflowgeneration,zheng2025mermaidflowredefiningagenticworkflow,shi2025flowagentachievingcomplianceflexibility,wang2025agentspeccustomizableruntimeenforcement,sharma2026contextcovderivingenforcingexecutable,openprose2026openprose,openclaw2026lobster}.
NLAHs share the premise that natural language can carry executable intent.
They differ in scope: the target object is the agent harness over a full task run, extending beyond an individual call, a fixed pipeline, or a formal workflow graph.
NLAHs deliberately keep a freer natural-language form to preserve editability and broad expressivity, while relying on IHR and deterministic hooks for execution.

\section{Conclusion}
We studied whether agent harness policy can be externalized as a compact, executable, and analyzable representation.
We introduced Natural-Language Agent Harnesses and an Intelligent Harness Runtime that executes these harnesses under shared runtime semantics.
Across coding, terminal-use, and computer-use benchmarks, IHR-executed NLAHs remain competitive with native harnesses while making the policy layer much shorter and easier to inspect.
Mechanism audits and module ablations further show that explicit harness documents can support process-level inspection and mechanism-level analysis.

\begin{ack}
We thank the reviewers for their careful reading and constructive feedback, which helped sharpen the scope, framing, and experimental design of this work.
We are also grateful to Ronak Malde and Thomas Wolf for valuable follow-up discussions.
\end{ack}

\bibliographystyle{plainnat}
\bibliography{custom,bib/benches_and_agents,bib/mcp,bib/harness_engineering,bib/related-work.exe-nl,bib/related-work.nl-and-workflow,bib/related-work.swarm-and-skill,bib/skill,bib/discussion,bib/skill_market,bib/agent,bib/agent_dynamic_orchestration}

\clearpage
\appendix
\crefalias{section}{appendix}

\section{Discussion}
\label{app:discussion}

\paragraph{NLAHs as a policy layer.}
The main lesson of this work is a division of labor.
Natural language is well suited for harness policy: roles, contracts, evidence requirements, retry rules, validation strategy, state handoff, and stopping conditions.
Code remains the right medium for exact mechanisms: parsers, tool execution, sandboxing, benchmark adapters, logging, and deterministic validators.
NLAHs make the policy layer explicit while IHR and deterministic hooks preserve executable precision.
This framing is narrower and more defensible than saying that natural language should replace controller code.

\paragraph{Why explicit harness policy matters.}
When harness logic is hidden inside a controller, researchers can usually compare whole systems but not cleanly compare the policy choices inside those systems.
NLAHs change the unit of analysis.
A harness can be read as text, executed under a shared runtime, audited through mechanism metrics, and modified module by module.
This makes harness design closer to an experimental object: we can ask what a verifier contributes, whether file-backed state matters, whether branching pays for its cost, and whether compression loses task-critical information.

\paragraph{Natural language remains useful at the harness level.}
Some prompt-level gains can diminish as models become stronger, and prompt tricks can be brittle across tasks \citep{wang2024advancedlanguagemodelseliminate,cao2024worstpromptperformancelarge}.
Our results point to a different role for natural language.
The useful object is an explicit run-level policy that tells an agent what evidence to preserve, when to delegate, how to verify, and when to stop.
This type of control remains important even when the base model improves, because stronger models still need task structure, state discipline, and acceptance criteria.

\paragraph{Toward harness representation science.}
Once harnesses become explicit objects, they become searchable and testable objects as well.
Future work can retrieve, compose, mutate, and optimize NLAH modules under shared runtime assumptions.
This creates a path from opaque harness engineering toward a more controlled science of harness representations, where the key questions include which full agent system wins and which harness policy choices cause the difference.

\section{Reproducibility}
\label{app:reproducibility}

The replication package, including our source code, is open-sourced at \url{https://github.com/curated-skills/LinguaClaw}.

\section{MHTBA Code-Artifact Portability on TB2}
\label{app:mhtba-portability}

The MHTBA code baseline is the released Meta-Harness TB2 artifact, whose public usage reports Terminal-Bench~2.0 results under \texttt{anthropic/claude-opus-4-6} with \texttt{--n-attempts 5}.
Our controlled RQ1 run intentionally places this same code artifact under the common GPT setting used in the paper, namely \texttt{gpt-5.4-mini}, reasoning effort \texttt{xhigh}, and one attempt.
This setting tests cross-model portability of a discovered harness artifact and avoids only remeasuring it in its native model environment.

\begin{table}[H]
\centering
\small
\caption{\textbf{MHTBA code-artifact timeout diagnostics on TB2 under GPT.}}
\label{tab:mhtba-code-timeout}
\begin{tabularx}{0.86\textwidth}{@{}lrrY@{}}
\toprule
Outcome group & Count & Share & Interpretation \\
\midrule
Resolved without timeout & 11 & 12.4\% & The code artifact both solved and stopped normally. \\
Resolved with timeout & 21 & 23.6\% & The verifier reward is already 1.0, but the agent loop did not terminate cleanly. \\
Failed without timeout & 12 & 13.5\% & The run failed for reasons other than the global agent timeout. \\
Failed with timeout & 45 & 50.6\% & The dominant failure mode is timeout after a long control loop. \\
\bottomrule
\end{tabularx}
\end{table}

\tabref{tab:mhtba-code-timeout} shows that the code-artifact gap is not simply a uniform inability to solve TB2 tasks.
Out of 89 samples, 66 end with \texttt{AgentTimeoutError}.
Among these timeout runs, 21 already have reward 1.0, so the task state satisfied the verifier but the controller still failed to stop.
The remaining 45 failed timeouts form the main source of the 32/89 code score.

\begin{table}[H]
\centering
\small
\caption{\textbf{MHTBA code-artifact trajectory diagnostics for cross-condition disagreements.}}
\label{tab:mhtba-disagreement-diagnostics}
\begin{tabularx}{0.98\textwidth}{@{}lrrrrY@{}}
\toprule
Slice & N & Timeouts & Mean episodes & Mean input tokens & Mean no-tool warnings \\
\midrule
Code fail, Prompt success & 26 & 25 & 316.5 & 16.3M & 146.9 \\
Code fail, NLAH success & 24 & 22 & 285.2 & 14.7M & 124.3 \\
Code fail, both Prompt and NLAH success & 19 & 19 & 318.3 & 15.4M & 146.4 \\
All three settings success & 23 & 16 & 174.3 & 5.2M & 79.1 \\
\bottomrule
\end{tabularx}
\end{table}

\tabref{tab:mhtba-disagreement-diagnostics} isolates the behavior pattern in cases where the GPT-based natural-language realizations succeed but the code artifact fails.
The code artifact usually spends hundreds of episodes in the agent loop, consumes tens of millions of prompt tokens, and accumulates many warnings that the previous response contained no tool calls.
This is a runtime-protocol symptom; task-domain difficulty explains only part of the pattern.

The trace-level mechanism is visible in the completion gate.
The code artifact first treats \texttt{task\_complete} as a pending-completion request and asks the model to call the completion tool again after a checklist.
If the next response does not call the completion tool again, the controller clears the pending-completion state and continues the loop.
Under GPT, many trajectories answer the confirmation prompt with text such as \texttt{DONE} and no tool call.
The controller then returns a no-tool warning, the model issues a harmless no-op command to satisfy the warning, and the run re-enters the first completion gate.
This loop can repeat until the global timeout.

The sample \texttt{tune-mjcf} is a compact example.
The verifier reward is 1.0, but the run still ends with \texttt{AgentTimeoutError} after 3600 seconds, 186 episodes, and 5.4M input tokens.
Near the end of the trace, GPT alternates between a \texttt{task\_complete} tool call, a text-only \texttt{DONE} response with \texttt{Tool Calls: []}, a no-tool warning, and a no-op shell command.
Thus the run has already reached a valid task state, but it cannot satisfy the code artifact's exact two-call stopping protocol.

This evidence supports a model-harness adaptation explanation.
The released artifact was optimized and reported in a Claude Opus 4.6 setting, and it includes Anthropic-specific prompt caching behavior.
When transplanted to GPT, assumptions about completion confirmation, tool-call selection, and long-loop stopping transfer poorly.
Prompt and NLAH keep the high-level terminal-harness ideas, such as environment bootstrapping, batched shell work, programmatic checking, and final review, while avoiding the brittle released state machine by running through the Codex execution substrate and writing reproducible artifacts such as \texttt{/sa-output/artifacts/solve.sh}.
In the inspected Prompt and NLAH TB2 trajectories, the run ends with the Codex task-complete event and does not enter a repeated no-tool warning cycle.
The MHTBA result therefore illustrates a general harness-engineering risk: a code harness discovered for one model can encode latent assumptions about that model's tool-calling and stopping behavior, and those assumptions may be more fragile than the natural-language harness policy it was meant to implement.

\section{Formalization and Definitions}

\subsection{Harness engineering aspects}
\label{app:harness-aspects}

\begin{table}[!htbp]
\centering
\caption{Harness-engineering aspects and their core design questions.}
\label{tab:harness-aspects}
\small
\setlength{\tabcolsep}{5pt}
\begin{tabularx}{\textwidth}{@{}p{0.28\textwidth}Y@{}}
\toprule
Aspect & Core question \\
\midrule
Agent loop & How does the agent observe, plan, act, validate, update state, retry, and stop over a long run? \\
Tool design and documentation & Which tools are exposed, how are their schemas and descriptions written, and how are tool errors returned? \\
Context engineering & What should each call see, what should be loaded just in time, and when should context be compressed or refreshed? \\
Filesystem and workspace & Where do task inputs, intermediate files, scratch files, and final artifacts live? \\
Memory and state & What information persists across steps, retries, context compression, or future runs, and what is the authoritative carrier? \\
Validation and stopping & What evidence is required before success, failure, retry, or budget exhaustion can be declared? \\
Safety, permissions, and sandboxing & What actions, files, network resources, commands, and credentials are allowed or blocked? \\
Runtime defaults & What default limits, timeouts, context policies, model settings, and execution modes shape behavior? \\
Observability, logging, and replay & What trace, provenance, artifacts, and metrics are recorded so a run can be audited or replayed? \\
Retry and recovery & How are failures classified, how is state restored, and what changes on the next attempt? \\
Budget control & How are token, time, tool-call, candidate, retry, and monetary budgets enforced or reported? \\
\bottomrule
\end{tabularx}
\end{table}

\subsection{NLAH expressivity boundary and code-harness mapping}
\label{app:nlah-expressivity}
\label{app:nlah-code-mapping}
\tabref{tab:nlah-code-boundary} specifies the boundary between natural language and code in an IHR+NLAH system.
The key boundary is whether an aspect should be shared substrate, shared runtime semantics, editable task-family policy, deterministic task-family hook, or model-internal behavior.

\begin{table}[!htbp]
\centering
\caption{Natural-language/code boundary in an IHR+NLAH system.}
\label{tab:nlah-code-boundary}
\small
\setlength{\tabcolsep}{4pt}
\begin{tabularx}{\textwidth}{@{}p{0.18\textwidth}p{0.16\textwidth}Y@{}}
\toprule
Layer & Owner & Responsibility \\
\midrule
Base runtime code & Code & Model APIs, LiteLLM routing, tool schemas, \texttt{bash} execution, timeouts, event streams, message history, context-token estimation, and run state; code also carries what must be precise, reproducible, safe, fast, or dependent on external systems, including tool calls, file execution, sandboxing, parsers, and evaluators. \\
Runtime policy / charter & Fixed NL & IHR interpretation semantics, including parent-child boundaries, the meaning of an agent call, artifact and \texttt{STATE\_ROOT} semantics, completion gates, and audit requirements; it also carries shared interpretation rules, such as the rule that a parent orchestrator does not do substantive task work while executor children perform the task. \\
NLAH & Replaceable NL & The concrete harness roles, stages, loops, validation policy, recovery policy, state strategy, delegation policy, and module composition; natural language should carry roles, stages, strategies, when to validate, how to recover, what evidence is sufficient, and which modules can be ablated. \\
Scripts / adapters & Code hook & Tests, validators, parsers, benchmark wrappers, artifact post-processing, and any operation that requires exact execution; script hooks are task-family-specific precise operations that should not rely on natural-language interpretation and are executed by agents from the filesystem through \texttt{bash}. \\
Model internals & Not NLAH & Constrained decoding, logit bias, sampling, and model-internal reasoning mechanisms, unless a runtime or provider exposes them as code-level hooks. \\
\bottomrule
\end{tabularx}
\end{table}

For production use, NLAH mainly offers fast iteration, auditing, portability, and module-level experimentation.
Its weaknesses are interpretation uncertainty, model dependence, cost, and safety or permission risks.
Production systems should therefore keep safety, permissions, evaluation, and key parsing logic in code.

\tabref{tab:code-to-nlah-mapping} maps the main harness-engineering aspects in \tabref{tab:harness-aspects} to the IHR+NLAH carriers.

\begin{table}[!htbp]
\centering
\caption{Mapping code-harness aspects to IHR+NLAH carriers.}
\label{tab:code-to-nlah-mapping}
\small
\setlength{\tabcolsep}{3pt}
\begin{tabularx}{\textwidth}{@{}p{0.22\textwidth}YYYY@{}}
\toprule
Aspect & Base-agent code & Runtime policy & NLAH & Scripts/adapters \\
\midrule
Agent loop & Minimal model-tool loop and child launch primitive & Parent orchestrator role, child boundary, stop discipline & Task-family stages, branching, role responsibilities & Optional loop drivers for fixed protocols \\
Tool design and documentation & Tool schemas, dispatch, timeout, error capture & Shared rules for unavailable tools and tool-result handling & When and why to use each tool class & Tool wrappers and capability adapters \\
Context engineering & Message history and context compression mechanism & Forking semantics, fresh-child semantics, handoff discipline & What to pass, reopen, summarize, or keep private & Retrieval or summarization helpers \\
Filesystem and workspace & Workspace access and run-state locations & Separation between runtime state and final artifacts & Required paths, artifact ownership, overwrite rules & Artifact processors and path normalizers \\
Memory and state & Minimal persisted run parameters, messages, and events & When durable state is required for audit or reuse & Task histories, manifests, memory files, state update rules & State serializers and merge scripts \\
Validation and stopping & Ability to run commands and collect outputs & Contract-first completion and failure reporting & Acceptance gates, verifier roles, evidence requirements & Tests, validators, parsers, graders \\
Safety, permissions, and sandboxing & Process execution limits and allowed tool surface & Shared permission interpretation and refusal discipline & Task-family prohibitions or safe-use rules & Sandbox wrappers and policy checks \\
Runtime defaults & Provider access, default timeout, max steps, model routing & Common execution defaults exposed to all NLAHs & Overrides only when task-family semantics require them & Configuration adapters \\
Observability, logging, and replay & Event stream, message log, tool records & What boundaries must remain inspectable & Claimed stages, module boundaries, required evidence notes & Trace post-processors and metrics scripts \\
Retry and recovery & Ability to rerun tools or launch new children & Failure labels, route changes, honest-stop semantics & Retry policy, repair stages, fallback paths & Cleanup, reset, and diagnostic scripts \\
Budget control & Token/time/tool-call accounting and caps & Shared budget discipline and reporting & Candidate counts, retry limits, search depth policy & Cost aggregation and budget-check scripts \\
\bottomrule
\end{tabularx}
\end{table}

This mapping defines the expressivity boundary used in this paper.
If a decision is shared by all harnesses or needed for machine execution, it belongs to base-agent code or runtime policy.
If a decision is task-family-specific and should be read, edited, refactored, or ablated, it belongs to NLAH.
If correctness depends on exact execution, it belongs to scripts or adapters.

\section{Runtime and Implementation Details}

\subsection{Runtime-policy prompt}
\label{app:runtime-policy-prompt}
The fixed runtime-policy prompt used in IHR encodes the shared cross-benchmark runtime policy that makes NLAHs executable under a common substrate.
In operational terms, the runtime-policy prompt enforces five ideas:
\begin{itemize}
\item \textbf{Runtime-only parent role.}
The top-level agent takes the orchestrator role, so even a nominally single-agent harness is realized as ``parent runtime + one task child.''
This keeps substantive workspace work inside child agents and makes delegation boundaries inspectable.
\item \textbf{Minimal delegated baseline.}
If no NLAH is provided, or if the provided NLAH is incomplete, the runtime first constructs the thinnest runnable baseline from the benchmark contract and then treats extra NLAH clauses as overlays on that baseline.
This baseline is part of what distinguishes IHR-executed NLAH from prompted NLAH: the runtime grounds task instructions in a runnable delegated execution substrate.
\item \textbf{Call-graph recovery with explicit context semantics.}
The runtime reconstructs roles, stages, repetition structure, and independence requirements from NLAH text, and then realizes them as child-agent launches.
\texttt{fork\_context=true} means that a child forks and inherits the parent's accumulated conversational context.
\texttt{fork\_context=false} means that a child starts from a fresh, independent, clean context and receives only the minimal task packet explicitly handed to it.
Together with disposable one-shot children and fresh children for independent branches, this preserves the original harness's model-call boundaries and avoids collapsing everything into one long dialogue.
\item \textbf{Separated runtime state and final artifacts.}
Durable intermediate state is written under \texttt{STATE\_ROOT} (default \texttt{/sa-output/runtime}) only when needed for reuse or auditability, while judgeable deliverables go to \texttt{/sa-output/artifacts}.
This lets the runtime expose stable evidence surfaces without mirroring the entire task workspace.
\item \textbf{Contract-first completion and auditability.}
Benchmark outputs and completion gates remain the primary contract, but the runtime must leave inspectable evidence when an NLAH claims staged or multi-role execution.
As a result, IHR-executed NLAH adds a shared layer of orchestration, context, artifact, and reporting discipline, with prompt text serving as one input to that layer.
\end{itemize}

\subsection{Realizing Harness Aspects with IHR}
\label{app:realizing-harness-aspects}

In IHR, an agent call is the atomic unit of harness execution.
A model call is a degenerate agent call in which the task instruction asks the agent to answer once without external action.
This agent-level view matches the level at which harnesses specify prompts, tools, workflow, memory, retrieval, compression, validation, and delegation.

Prompt design is realized by constructing the initial context of each agent.
When a harness requires role-specific system prompts, the NLAH can store each prompt in a separate file, and IHR can instruct the corresponding agent instance to load that file.
This gives the NLAH a relatively precise way to specify agent roles while still keeping prompt content in editable natural language.

Tool design is realized through code-backed tools.
Concrete tools are executable programs, wrappers, scripts, services, or adapters.
An agent with terminal access can call these tools by running the referenced code under the sandbox, permission, and budget constraints of the run.
Thus, NLAH carries tool policy and tool-use discipline, while scripts and adapters carry exact tool behavior.

Workflow and multi-agent structure are realized by IHR orchestration.
IHR can launch new agents, send messages to running agents, inspect their returned state or answers, route outputs to other agents, launch additional agents, and close agents whose roles are complete.
This lets natural-language harness policy materialize as concrete call boundaries, handoffs, verification stages, selector stages, and multi-agent execution.

Memory and retrieval are realized as external, path-addressable state.
The NLAH specifies which facts, decisions, failures, validation results, and artifacts should be written to memory files, task histories, manifests, or evidence records.
IHR can require later agents to reopen these files before planning, verification, handoff, or final reporting.
Retrieval follows the same pattern: the NLAH specifies retrieval sources, timing, and evidence rules, while IHR exposes search scripts, file indexes, or adapters through the terminal and requires retrieved evidence to be written into auditable artifacts.

Compression is realized as a contract-preserving context operation.
The NLAH specifies when compression is allowed, what state must be externalized before compression, what fields the compressed state must preserve, and which information must remain recoverable.
When context becomes long or a stage boundary is reached, IHR can require the agent to write a compact state file containing the task goal, constraints, explored paths, failure signals, validation status, key evidence, artifact paths, and next actions.
Later agents recover state by reopening this compressed file and the referenced artifacts.
This makes compression an auditable harness operation with explicit recovery surfaces.

\section{NLAH Modules}
\label{app:nlah-modules}
The module boxes below are concise paraphrastic summaries of the NLAH module behaviors used in RQ3.

\begin{rqmodulebox}{file-backed state}
ROOT: Choose STATE_ROOT under /sa-output, keep it separate from the original task workspace, and maintain STATE_ROOT/RESPONSE.md as the stable runtime-level status file.
HANDOFF: No prompt, role instruction, reply, or promoted artifact counts as transferred until it exists as TASK.md, NLAH.md, RESPONSE.md, or another named file under STATE_ROOT.
CHILD PACKET: Each launched child receives children/<id>/TASK.md, optional children/<id>/NLAH.md, and writes back children/<id>/RESPONSE.md.
BOOKKEEPING: Keep append-only launch and promotion history in state/task_history.jsonl, index promoted outputs in artifacts/manifest.json, and reopen files by path for reuse and recovery.
\end{rqmodulebox}

\begin{rqmodulebox}{evidence-backed answering}
ARTIFACT: Before any final answer, final patch, or solved claim, write one standalone evidence document as the designated evidence artifact for the current task or stage.
STRUCTURE: Cover the problem statement, relevant materials, observed symptoms, root cause, candidate resolution, validation, and residual uncertainty.
CLAIM DISCIPLINE: Each major claim must state its provenance, whether it is direct observation or inference, and the minimal supporting span or output segment when available.
GATE: Do not release a complete answer while release-critical claims remain uncited, contradicted, or materially incomplete in that evidence document.
\end{rqmodulebox}

\begin{rqmodulebox}{verifier separation}
ROLE: Verifier inspects one candidate answer against the original problem and the lightest sufficient task materials needed to check it.
PROCEDURE: Identify the candidate's claim, break it into checkable subclaims, audit completeness, factual correctness, and logical correctness, and run at least one central independent check when feasible.
OUTPUT: Return exactly one primary verdict label plus a report that explains the verdict, names the checks run or blocked, and does not repair the candidate on its behalf.
\end{rqmodulebox}

\begin{rqmodulebox}{self-evolution}
LOOP: Run an explicit retry loop with a real baseline attempt first and a default cap of five attempts unless the task specifies otherwise.
TRIGGER: After every non-successful, partially successful, unstable, or stalled attempt, reflect on concrete failure signals before planning the next attempt.
AXES: Redesign the next attempt along prompt, tool, and workflow evolution, and make attempt 2 materially reflect the reflection from attempt 1.
STOP: Continue until judged success or the attempt cap is reached, and report incomplete to avoid pretending the last attempt passed.
\end{rqmodulebox}

\begin{rqmodulebox}{multi-candidate search}
BUDGET: Use an explicit candidate budget K, defaulting to K=5 when unspecified, and restore lost budget if a branch crashes before returning comparable evidence.
DIVERSITY: Vary the core hypothesis, decomposition, evidence route, tool plan, or risk preference so candidates are not near-duplicates.
SELECTION: Prune duplicates, unsupported, dominated, or overly risky branches, then compare survivors on task fit, evidence quality, coherence, and repair cost.
ESCALATION: If no candidate is good enough, expand or redesign the search and avoid forcing a fragile winner.
\end{rqmodulebox}

\begin{rqmodulebox}{dynamic orchestration}
AUTONOMY: Beyond the mandatory task-owning child, add extra subagents only when delegation materially improves coverage, latency, specialist focus, or quality control, and prefer the smallest adequate topology.
TOPOLOGY: Classify the task shape, assign each child a non-overlapping responsibility and success condition, and parallelize only genuinely independent branches.
PARENT ROLE: Once a delegated topology is chosen, the parent should narrate launches, waits, comparisons, and integration while leaving substantive work to the corresponding child roles.
BOUNDARY: Direct task-workspace familiarization or repository probing belongs to child roles after commitment to delegated execution.
\end{rqmodulebox}

\begin{rqmodulebox}{context compression}
TRIGGER: Compress active context only after writing or reopening path-addressable state that preserves the current task contract, artifacts, failures, and next actions.
SUMMARY: Preserve the task goal, constraints, explored paths, accepted and rejected decisions, validation status, unresolved risks, and exact artifact paths.
FILTER: Drop redundant dialogue, stale speculation, and low-value logs, but never drop acceptance criteria, error signatures, commands still needed for replay, or handoff state.
CHECK: After compression, reload the compacted state and continue only after confirming that the next action and required evidence remain recoverable.
\end{rqmodulebox}

\begin{rqmodulebox}{markdown memory}
CARRIER: Maintain a Markdown memory file with stable headings for task facts, decisions, reusable observations, environment notes, and unresolved caveats.
UPDATE: Add concise entries after meaningful discoveries, validations, failures, or design decisions, and mark superseded entries to avoid silently overwriting them.
READ: Reopen the memory file before planning, delegation, verification, and final reporting so durable facts can re-enter active context by path.
HYGIENE: Keep entries source-grounded and task-scoped; avoid turning memory into a transcript or a dumping ground for speculative notes.
\end{rqmodulebox}

\subsection{Additional RQ3 module observations}
\label{app:rq3-additional}

\paragraph{Validation helps, but only when it stays close to the benchmark gate.}
\emph{Verifier} is positive on both benchmarks, but its gains are uneven: +0.2 on SWE and +8.4 on OSWorld.
This is still a useful result.
It shows that an explicit checking stage can help, but only when the verifier's local object of judgment remains close to the benchmark's own acceptance criterion.
The same logic explains why \emph{dynamic orchestration} is real but modest: it changes behavior enough to recover some cases, yet its aggregate gains remain limited because it does not automatically tighten the final acceptance path.

\paragraph{Durable state is more reliable than aggressive compression.}
The memory-related modules separate cleanly.
\emph{Context compression} hurts both benchmarks, dropping SWE from 73.0 to 72.0 and OSWorld from 44.4 to 36.1.
\emph{Markdown memory} is mixed, hurting SWE by 2.8 points but helping OSWorld by 5.6.
By contrast, \emph{file-backed state} is positive in both settings.
The interpretation is that path-addressable durable state is a stronger and more reliable carrier than either aggressive summarization or free-form memory notes, especially when the evaluator depends on action-critical details that are easy to lose during compression.

\section{Limitations, Risks, and Broader Impact}

\subsection*{Limitations}
The main limitation is natural-language imprecision.
NLAHs are editable natural-language policies, so semantically important constraints may be under-specified, interpreted differently across models, or weakened by paraphrase.
We therefore keep exact mechanisms in code and treat executed behavior as something that must be checked through runs, avoiding inference from the text alone.

\subsection*{Broader impact and risks}
Externalizing harness modules can reduce development cost, improve comparability, and encourage reuse of robust workflows.
However, portable harness logic and scripts may also lower the barrier to spreading risky workflows.
Because harnesses mediate tool use, artifact handling, and delegation, they can introduce new attack surfaces for prompt injection, malicious tool grafting, or supply-chain contamination.
Deployments should combine provenance tracking, review, permission control, and sandbox isolation.


\end{document}